\documentclass[runningheads]{llncs}
\usepackage{amsmath}
\usepackage{algorithm}
\usepackage{algpseudocode}
\usepackage{graphicx}
\usepackage{amssymb}
\usepackage[T1]{fontenc}
\usepackage{array}
\usepackage{booktabs}
\usepackage{multirow}
\usepackage{graphicx,verbatim}
\usepackage{svg}
\usepackage{hyperref}
\usepackage{color}

\usepackage{makecell} 

\usepackage{threeparttable} 

\begin{document}
\title{SemiVT-Surge: Semi-Supervised Video Transformer for Surgical Phase Recognition}
\titlerunning{Semi-Supervised Video Transformer for Surgical Phase Recognition}
%
\author{Yiping Li\inst{1}
\and Ronald de Jong\inst{1}
\and Sahar Nasirihaghighi\inst{2}
\and Tim Jaspers\inst{3}
\and Romy van Jaarsveld\inst{4}
\and Gino Kuiper\inst{4}
\and Richard van Hillegersberg\inst{4}
\and Fons van der Sommen\inst{3}
\and Jelle Ruurda\inst{4}
\and Marcel Breeuwer\inst{1}
\and Yasmina Al Khalil\inst{1}
}
\authorrunning{Y. Li et al.}
%
\institute{
Department of Biomedical Engineering, Medical Image Analysis, Eindhoven University of Technology, Eindhoven, The Netherlands
\and
Institute of Information Technology (ITEC), University of Klagenfurt, Austria
\and
Department of Electrical Engineering, Video Coding \& Architectures, Eindhoven University of Technology, Eindhoven, The Netherlands
\and
Department of Surgery, University Medical Center Utrecht, Utrecht, The Netherlands
}



\maketitle              
\begin{abstract}
Accurate surgical phase recognition is crucial for computer-assisted interventions and surgical video analysis. Annotating long surgical videos is labor-intensive, driving research toward leveraging unlabeled data for strong performance with minimal annotations. Although self-supervised learning has gained popularity by enabling large-scale pretraining followed by fine-tuning on small labeled subsets, semi-supervised approaches remain largely underexplored in the surgical domain. In this work, we propose a video transformer-based model with a robust pseudo-labeling framework. Our method incorporates temporal consistency regularization for unlabeled data and contrastive learning with class prototypes, which leverages both labeled data and pseudo-labels to refine the feature space. Through extensive experiments on the private RAMIE (Robot-Assisted Minimally Invasive Esophagectomy) dataset and the public Cholec80 dataset, we demonstrate the effectiveness of our approach. By incorporating unlabeled data, we achieve state-of-the-art performance on RAMIE with a 4.9\% accuracy increase and obtain comparable results to full supervision while using only 1/4 of the labeled data on Cholec80. Our findings establish a strong benchmark for semi-supervised surgical phase recognition, paving the way for future research in this domain. Code is available at \href{https://github.com/IntraSurge/SemiVT-Surge}{https://github.com/IntraSurge/SemiVT-Surge}.


\end{abstract}
\section{Introduction}
Surgical phase recognition is used in computer-assisted surgery systems to classify each frame from surgical video footage into different stages of a surgical procedure. It supports context-aware assistance and decision support, enhances workflow efficiency, facilitates postoperative analysis, surgeon performance evaluation, and identification of problematic phases \cite{maier2022surgical}. It remains a high-level understanding task, presenting significant challenges as the model must distinguish between similar semantics across different phases of a procedure.

Transformers have become increasingly popular in surgical phase recognition, with tailored attention mechanisms to enhance spatio-temporal modeling. SKiT \cite{liu2023skit} introduced critical pooling to record key information and self- and cross-attention for feature fusion.  Label-Guided Teacher \cite{guan2024label} incorporated labels as extra supervision via cross-attention to refine feature representations. To better capture temporal dynamics of video data, researchers have recently shifted to video-based models, instead of relying on per-frame feature extraction. Surgformer \cite{yang2024surgformer} enhanced TimeSformer \cite{bertasius2021space} with Hierarchical Temporal Attention to improve spatio-temporal feature capture, while MuST \cite{perez2024must} proposed a multi-scale Transformer that captures short-, mid-, and long-term dependencies.

Annotating surgical phases is labor-intensive, requiring frame-by-frame review. To reduce supervision, self-supervised learning has gained traction, leveraging large surgical datasets for pretraining before fine-tuning on downstream tasks such as phase recognition and semantic segmentation \cite{alapatt2024jumpstarting,batic2024endovit,jaspers2024exploring,jaspers2025scaling} , demonstrating promising results. Surgery-specific pretraining has been shown to improve performance, whereas cross-surgery pretraining may even underperform compared to models trained on general computer vision datasets \cite{alapatt2024jumpstarting}. Given the procedure-specific nature of surgical phases, semi-supervised learning offers a promising way to leverage unlabeled data from the same surgery and reduce annotation effort. However, it remains underexplored in surgical phase recognition, with only a few studies addressing its potential. SurgSSL \cite{shi2021semi} introduced a two-stage semi-supervised approach with visual and temporal consistency as regularization and pseudo-labeling for refinement, while FedCy \cite{kassem2022federated} applied semi-supervised learning in a federated setting using temporal cycle consistency and contrastive learning for multi-center data. Both methods, developed early on, were based on simple convolutional neural networks.

Beyond surgical applications, semi-supervised learning has been widely explored in image and short-term video tasks. FixMatch \cite{sohn2020fixmatch} introduced consistency regularization with weak and strong augmentations for image classification, while Mean Teacher \cite{tarvainen2017mean} demonstrated the effectiveness of an exponential moving average (EMA) strategy for semi-supervised learning. More recent advances \cite{dave2023timebalance,xing2023svformer,xu2022cross} have further refined augmentation strategies and temporal modeling for short-term action recognition tasks, yet their applicability to long-duration tasks like surgical phase recognition remains unexplored.
Contrastive learning is well-established for learning discriminative representations. In supervised settings, it has proven effective for surgical phase recognition \cite{guan2024label}. In semi-supervised medical image classification and segmentation, class prototypes have been used to structure the feature space around class-specific embeddings \cite{basak2023pseudo,he2024open}, improving learning with pseudo-label guidance, which inspired our work.

\begin{figure}[ht!]
\label{overview}
\centering
\includegraphics[width=\textwidth]{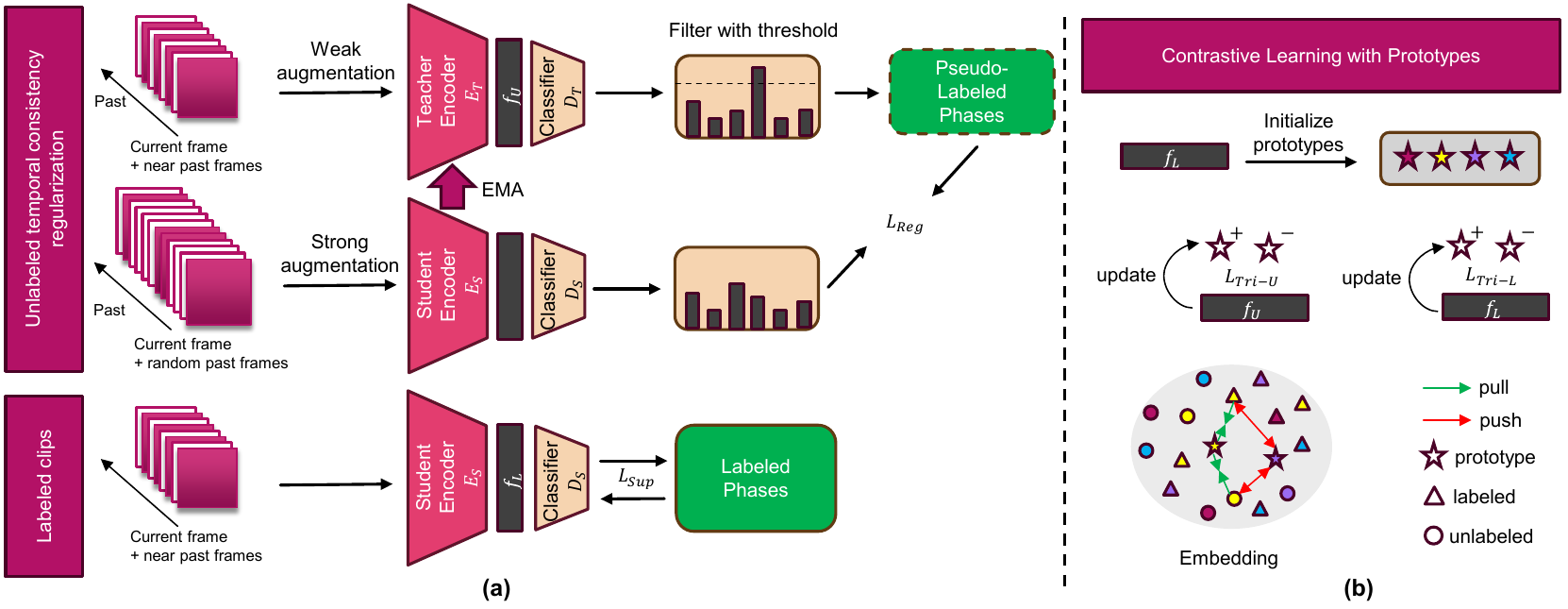}
\caption{Overview of the proposed method: (a) overall training process with temporal consistency regularization, as detailed in Section \ref{section: temporal consistency}, and (b) contrastive learning with class prototypes, as detailed in Section \ref{section: contrastive}.}
\label{fig1}
\end{figure}

In this work, we explore the potential of semi-supervised learning for surgical phase recognition using a video transformer model. An overview of our approach is provided in Fig. \ref{overview}. The training procedure involves two key components: (a) long- and short-term frame sampling for weak and strong augmentations during pseudo-labeling to improve temporal consistency; and (b) contrastive learning with class prototypes to regularize the feature space using both labeled and pseudo-labeled data, with prototypes continuously updated to refine feature embeddings. We demonstrate the superiority of our method over state-of-the-art approaches through comprehensive experiments on two datasets.

\section{Methods}
\subsection{Overview of the training strategy}
In this work, we address the problem of online surgical phase recognition, a video classification task that predicts the surgical phase \( Y_t \) of the current frame from video input \( I_t \). Given a video stream \( I_t = \{I_j\}_{j=1}^t \) up to time \( t \), we aim to learn a mapping \( M_\theta \) such that \( M_\theta(I_t) \approx Y_t \), where \( Y_t \in \{1, \dots, C\} \) represents the surgical phase at time \( t \) and \( C \) is the total number of phases. We use TimeSformer \cite{bertasius2021space} as the video encoder \( E \), which extracts spatiotemporal features \( f_t = E(I_t) \). Here, \( f_t \) is the class token with a feature embedding size of 768. The input consists of \( T=16 \) frames, including the current frame and 15 past frames. A classifier \( D \), implemented as a linear layer, predicts the phase probability distribution \( Y_t = D(f_t) \). \( \theta \) denotes model parameters. The success of this task relies on training a robust encoder \( E \) for high-quality feature representations. 
The training begins with a warm-up on labeled data using a standard cross-entropy loss \( L_{\text{Sup}} \), followed by semi-supervised training with both labeled and unlabeled data, where the supervised loss \( L_{\text{Sup}} \) is still applied to the labeled data, while additional losses are introduced. Specifically, the temporal consistency loss \( L_{\text{Reg}} \) is applied solely to the unlabeled data (see Section \ref{section: temporal consistency} for details). The contrastive losses \( L_{\text{Tri-U}} \) and \( L_{\text{Tri-L}} \) regularize the feature embeddings using updated class prototypes with both labeled and unlabeled data (see Section \ref{section: contrastive} for details). 
The total loss  \( L_{\text{total}} \) is the sum of these individual losses. For a detailed description of the full workflow, refer to Algorithm \ref{alg:workflow}.

\begin{algorithm}[ht!]
\small
\caption{Workflow of Our Proposed Method}
\label{alg:workflow}
\begin{algorithmic}
\Require Labeled dataset $\mathcal{S}$, unlabeled dataset $\mathcal{U}$
\State \textbf{Warm-up:} Train \(E_S, D_S\) with \(L_{\text{Sup}}\) to initialize \(\theta_{E,S}, \theta_{D,S}\).  
\State \textbf{Teacher Init:} Set \(\theta_{E,T} \gets \theta_{E,S}, \theta_{D,T} \gets \theta_{D,S}\).  
\State \textbf{Prototypes Initialization:} mean features from labeled dataset: $\mathcal{C} \gets \{\mu_c\}_{c=1}^C$ where $\mu_c = \frac{1}{|I_L^c|}\sum Normalize(E_S(I_L^c))$ 

\While{iteration $\leq$ max iteration}
    \State Sample mini-batches \( B_L \) from \( \mathcal{S} \) and \( B_U \) from \( \mathcal{U} \)
    
    \For{$(I_L, Y_L) \in B_L$}
        \State $Y'_L \gets D_S(E_S(I_L))$ \quad \Comment{Student prediction}
        \State $L_{\text{Sup}} \gets \text{CE}(Y'_L, Y_L)$ \quad \Comment{Supervised loss}
        \State $f_L \gets \text{Normalize}(E_S(I_L))$ \quad \Comment{Feature embedding}
        \State $L_{\text{Tri-L}} \gets \text{TripletLoss}(f_L, \mathcal{C}_{Y_L}, \{\mathcal{C}_k\}_{k \neq Y_L})$ 
        \State Update prototypes: $\mathcal{C}_{Y_L} \gets \eta\mathcal{C}_{Y_L} + (1-\eta)f_L$ 
    \EndFor
    \For{$I_U \in B_U$}  
        \State $I^s_U \gets \text{StrongAugment}(I_U)$ \Comment{Strong augmentation}
        \State $I^w_U \gets \text{WeakAugment}(I_U)$ \Comment{Weak augmentation}
        \State $Y'_{U,S} \gets D_S(E_S(I^s_U))$ \Comment{Student prediction}
        \State $Y'_{U,T} \gets D_T(E_T(I^w_U))$ \Comment{Teacher prediction}
        \State $f_U \gets \text{Normalize}(E_T(I^w_U))$ \quad \Comment{Feature embedding}
        \State Filter $\mathcal{F} \gets \{(f_U, Y'_{U,S}, Y'_{U,T}) | \max(Y'_{U,T}) \geq \delta\}$ \Comment{High-confidence subset only}
        \For{$(f_U, Y_S, Y_T) \in \mathcal{F}$}
            \State $L_{\text{Tri-U}} \gets \text{TripletLoss}(f_U, \mathcal{C}_{\arg\max Y_T}, \{\mathcal{C}_k\}_{k \neq \arg\max Y_T})$
            \State $L_{\text{Reg}} \gets \text{CE}(Y_S, Y_T)$ \Comment{Temporal Consistency Regularization}
            \State Update prototypes: $\mathcal{C}_{Y_T} \gets \eta\mathcal{C}_{Y_T} + (1-\eta)f_U$ 
        \EndFor
    \EndFor

    \State Aggregate losses: $L_{\text{total}} \gets L_{\text{Sup}} + L_{\text{Reg}} + L_{\text{Tri-U}} + L_{\text{Tri-L}}$

    \State Update student: $\theta_{E,S}, \theta_{D,S} \gets \theta - \nabla_{\theta} L_{\text{total}}$
    \State Update teacher: $\theta_{E,T}, \theta_{D,T} \gets \alpha\theta_T + (1-\alpha)\theta_S$ \Comment{EMA}

\EndWhile
\State \Return $\theta_{E,T}, \theta_{D,T}$
\end{algorithmic}
\end{algorithm}

\subsection{Temporal Consistency Regularization}
\label{section: temporal consistency}
Our semi-supervised learning approach enforces consistency between differently augmented views of unlabeled data using a teacher-student architecture. The student model is shared across both labeled and unlabeled batches, and during training, the student model receives updates via gradients, while the teacher model is updated using EMA. Considering the nature of surgical phase recognition, temporal sampling itself serves as a natural augmentation strategy, enabling the model to learn robust representations across different temporal contexts. Given an unlabeled batch \( B_U \), each sample \( I_U \in B_U \) undergoes two augmentations: a weakly augmented view \( I^w_U = \text{WeakAugment}(I_U) \), which preserves short-term temporal context by including the processing frame and its preceding \( T-1 \) consecutive frames, and a strongly augmented view \( I^s_U = \text{StrongAugment}(I_U) \), which enforces long-term temporal alignment by randomly selecting \( T-1 \) frames from the full video history. Further details on the augmentation process can be found in Section \ref{implementation details}. The student model predicts \( Y'_{U,S} = D_S(E_S(I^s_U)) \), while the teacher model, predicts \( Y'_{U,T} = D_T(E_T(I^w_U)) \). To ensure reliable pseudo-labeling, we introduce a confidence-based selection mechanism: only samples where the maximum predicted class probability exceeds a predefined threshold \( \delta \) contribute to the unsupervised loss. Formally, the loss is defined as  

\begin{equation}
\label{cross entropy loss with threshold}
\mathcal{L}_{U} = \frac{1}{N_U} \sum_{I_U \in B_U} \mathbb{I} \left( \max(Y'_{U,T}) > \delta \right) \text{CE}(Y'_{U,S}, Y'_{U,T}),
\end{equation}

where \( \mathbb{I}(\cdot) \) is the indicator function that equals 1 when the confidence threshold is met, filtering out uncertain pseudo-labels, and \( \text{CE} \) denotes the standard cross-entropy loss. This dual-stream approach, coupled with confidence-based selection, facilitates effective feature learning while reducing the influence of noisy pseudo-labels.

\subsection{Contrastive Learning with Prototypes}
\label{section: contrastive}
Our method integrates semi-supervised learning with contrastive prototype alignment to regularize the feature embedding space. We employ triplet margin loss, computed in Euclidean space, to structure the embedding space by enforcing a margin between positive and negative samples. Normalizing the features ensures that distance measurements remain consistent and robust, thereby enhancing class separability. Class-specific prototypes are initialized as cluster centroids in a normalized feature space computed from labeled data with warmed-up model. These prototypes evolve through EMA updates during training. Unlabeled samples contribute only when teacher-generated pseudo-labels exceed a confidence threshold \( \delta \), filtering out unreliable labels as described in Equation \ref{cross entropy loss with threshold}.  

The prototype-anchored triplet loss enforces discriminative feature learning by pulling samples toward their class prototype while repelling them from negative prototypes:

\begin{equation}
\mathcal{L}_{\text{tri}} = \max\left(d(f, \mathcal{C}_y) - d\left(f, \frac{1}{k} \sum_{i=1}^{k} \mathcal{C}_{\text{neg}, i} \right) + m, 0\right),
\end{equation}

where \( d(\cdot, \cdot) \) denotes the Euclidean distance, \( m \) is the margin (empirically set to 0.3), and the negative prototype \( \mathcal{C}_{\text{neg}} \) is computed as the average of the \( k \) hardest negative prototypes, with \( k=3 \) selected as the nearest negative prototypes to the feature \( f \). For labeled data, positive and negative prototypes are selected based on the ground truth labels, while for unlabeled data, they are determined according to the pseudo-labels generated by the teacher model \( Y_T \).

\subsection{Implementation details}
\label{implementation details}
We use the TimeSformer \cite{bertasius2021space} model, initialized with pre-trained weights from the Kinetics-400 dataset \cite{kay2017kinetics}, which has been shown to accelerate convergence. Video frames are pre-extracted at 1fps, and resized to 256 \(\times\) 256. We first warm up with training only on labeled data for 3 epochs, and perform semi-supervised training for another 12 epochs. Data augmentation is performed using the AutoAugment library \cite{cubuk2019autoaugment}, with a strong augmentation configuration of 'rand-m9-n5-mstd0.8-inc1'. For weak augmentation, we apply random cropping, normalization, random rotation, and random crop. All models are trained with a batch size 16 on a single NVIDIA H100 GPU, with a SGD optimizer using a momentum of 0.9 and a weight decay of 0.001. For each setting, the basic learning rate is set to 0.005, and halved at epoch 8 and 12. The confidence threshold \( \delta \) was emperically chosen as 0.6 for RAMIE dataset and 0.8 for Cholec80. The EMA update rules are detailed in Algorithm \ref{alg:workflow}, with the EMA parameters for updating both the teacher model and prototypes set to \(\alpha =\eta = 0.9 \).

\section{Experiments and results}
\subsubsection{Datasets.}
Cholec80 \cite{twinanda2016endonet}: This dataset consists of 80 cholecystectomy surgery videos, each annotated with 7 distinct surgical phases. The dataset is divided into official training and test sets, with each containing 40 videos. For experiments with reduced annotated data, we randomly sample a subset of the training set as labeled data, while treating the remaining videos as unlabeled.

RAMIE \cite{anonymous}: The RAMIE dataset is a private dataset consisting of 27 labeled Robot-Assisted Minimally Invasive Esophagectomy videos. The dataset is split into 14 videos for training, 4 for validation, and 9 for testing, with annotations for 13 surgical phases. Compared to cholecystectomy, RAMIE surgery exhibits more complex temporal dynamics, with phase repetitions and increased variability in phase sequences. The unlabeled data is drawn from the same collection, with 20 videos randomly selected from a pool of 70 available unlabeled videos.

\subsubsection{Evaluation details and metrics.}
Our model is evaluated in an online manner, sliding through the video with a window size of \( T=16 \) to generate frame-wise predictions. The evaluation metrics include widely-used benchmark metrics: video-level Accuracy, phase-level Precision, phase-level Recall, phase-level Jaccard, and phase-level F1 score, implemented with the code from \cite{funke2023metrics}. Mean ± standard deviation is computed across videos in the test set.

\subsubsection{Ablation studies.}

\begin{table}[htbp!]
\centering
\caption{Ablative testing results on RAMIE and Cholec80 datasets. Sup. indicates supervised training with labeled data only. TCR represents the use of Temporal Consistency Regularization, CLP represents the inclusion of Contrastive Learning with Prototypes, and TCN represents the addition of TeCNO's causal TCN. A checkmark (\checkmark) indicates the presence of an attribute, while (-) denotes its absence.}
\label{tab:ablation_study}
\resizebox{\textwidth}{!}{%
\begin{tabular}{cccccccccccc}
\hline
\multirow{2}{*}{Sup.} & \multirow{2}{*}{TCR} & \multirow{2}{*}{CLP} & \multirow{2}{*}{TCN} & 
\multicolumn{4}{c}{RAMIE} & \multicolumn{4}{c}{Cholec80 (20 labeled training) } \\ 
\cline{5-12}
 & & & & Accuracy & Precision & Recall & Jaccard & Accuracy & Precision & Recall & Jaccard \\
\hline
\checkmark & - & - & -  & 78.7 ± 4.2 & 79.7 ± 4.0 & 74.7 ± 5.2 & 60.9 ± 4.3 & 84.3 ± 10.7 & 81.4 ± 9.0 & 79.0 ± 9.4 & 65.8 ± 13.2  \\
\checkmark & \checkmark & - & - & 80.4 ± 3.3 & 79.8 ± 3.4 & 75.6 ± 4.9 & 62.2 ± 3.9  &  87.0 ± 6.6 & 83.7 ± 8.3 & 82.9 ± 6.8 & 70.7 ± 10.2  \\
\checkmark & \checkmark & \checkmark & - & 81.9 ± 2.9 & 80.6 ± 3.8 & 78.9 ± 3.7 & 64.9 ± 3.4  & 89.6 ± 7.0 & 86.5 ± 8.0 & 86.5 ± 6.3 & 75.9 ± 10.7  \\
\checkmark & \checkmark & \checkmark & \checkmark & \textbf{83.4 ± 3.7} & \textbf{81.6 ± 3.4} & \textbf{78.8 ± 3.8} & \textbf{66.1 ± 3.1} & \textbf{90.4 ± 7.0} & \textbf{88.4 ± 7.2} & \textbf{86.5 ± 7.4} & \textbf{77.5 ± 11.3} \\

\hline
\end{tabular}%
}
\end{table}
We conduct ablation studies to assess the effectiveness of each component during training. Among these, the results for supervised learning and Temporal Consistency Regularization (TCR) can be compared to the popular semi-supervised learning method, FixMatch \cite{sohn2020fixmatch}, which highlights the effectiveness of our temporal regularization. 

With the addition of Contrastive Learning with Prototypes (CLP), we observe a more significant improvement. Since surgical phase recognition is essentially a classification task, introducing greater contrast between classes proves to be highly beneficial. This serves as a regularization method within the embedding space.

\begin{figure}[htbp!]
    \centering
    \includegraphics[height=2.8cm]{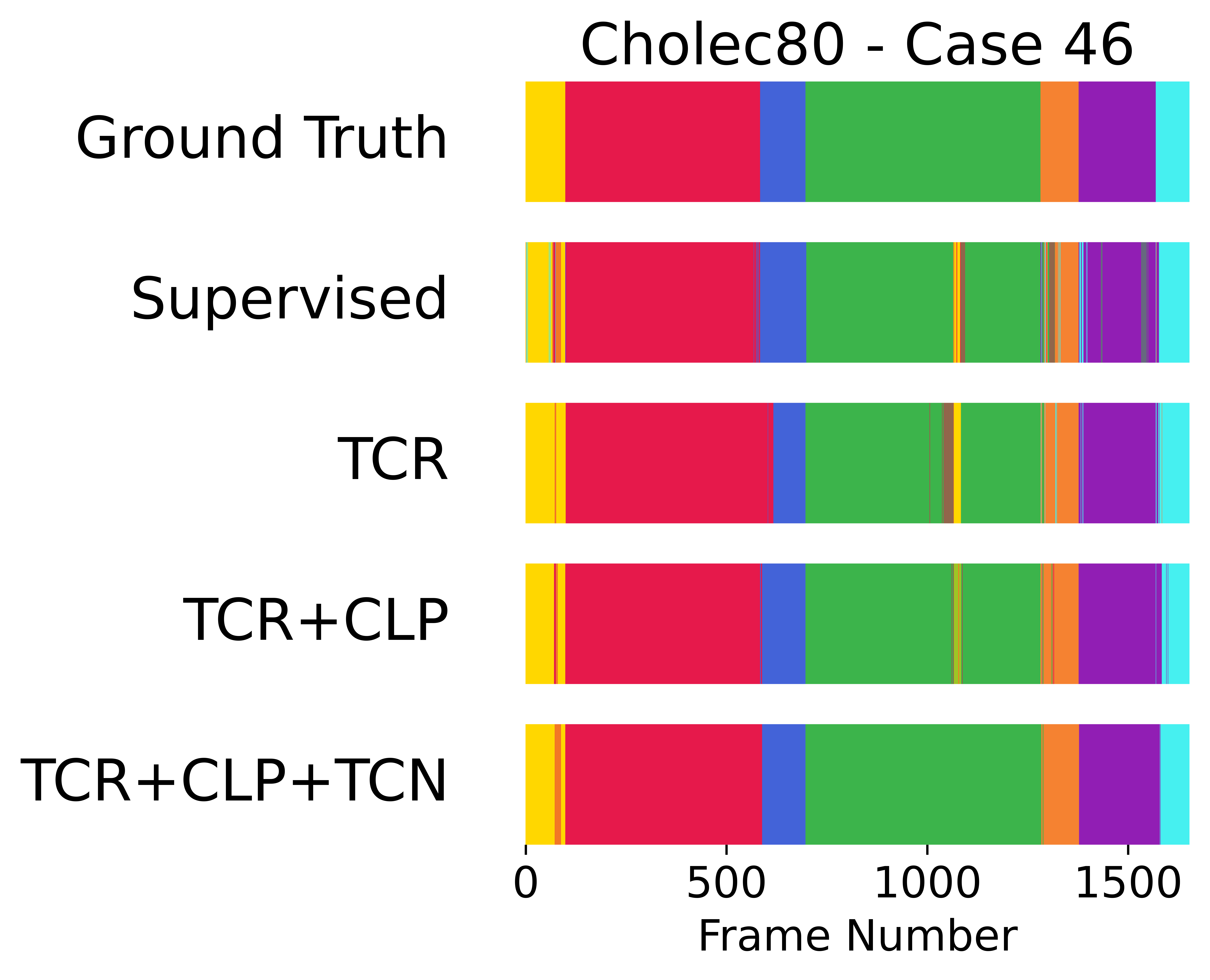}
    \hspace{0.05\textwidth}
    \includegraphics[height=2.8cm]{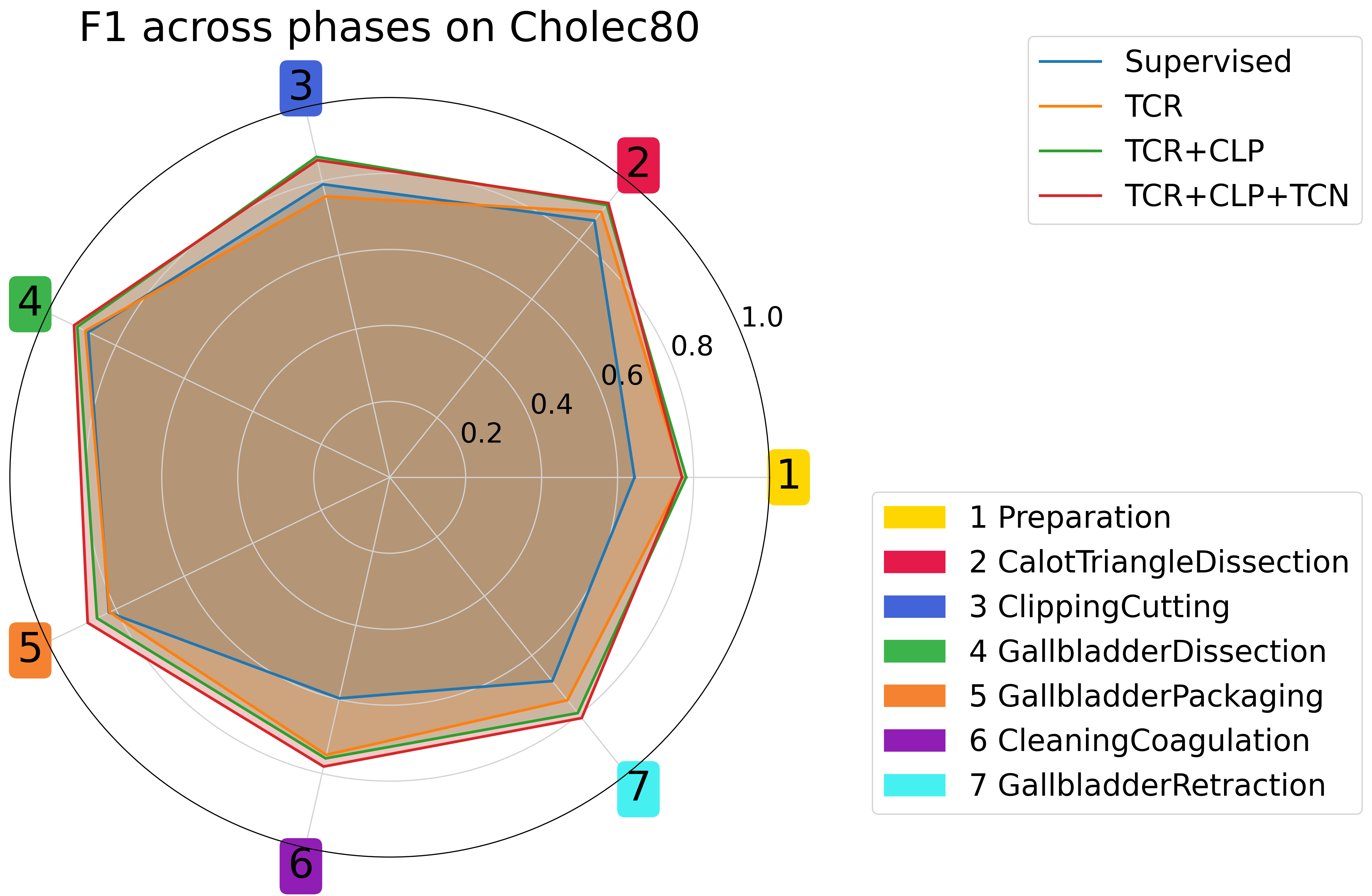}
    \caption{Cholec80: Qualitative result (left) and mean F1 scores across phases (right).}
    \label{fig:Cholec80}
\end{figure}

\begin{figure}[htbp!]
    \centering
    \includegraphics[height=2.8cm]{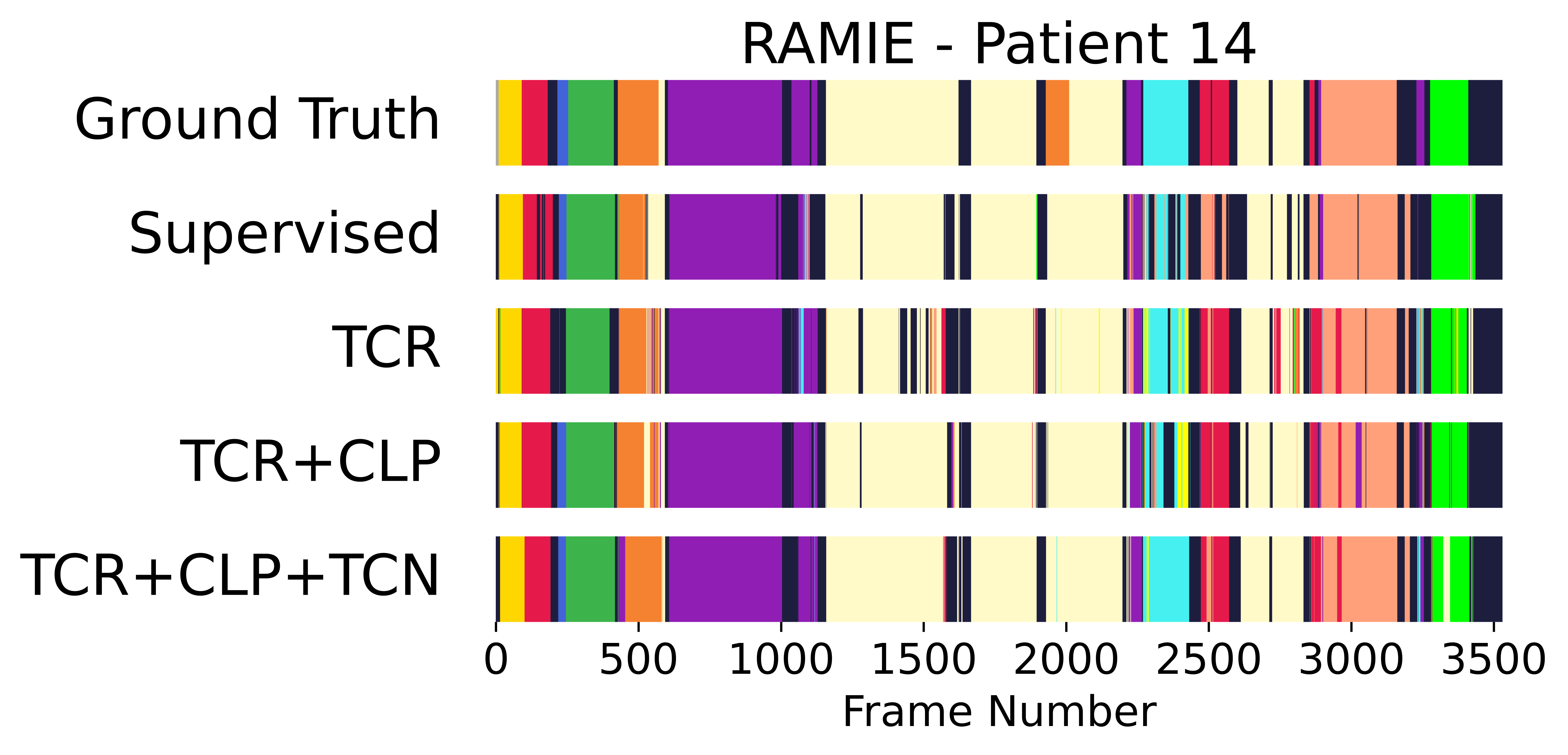}
    \hspace{0.0001\textwidth}
    \includegraphics[height=3.2cm]{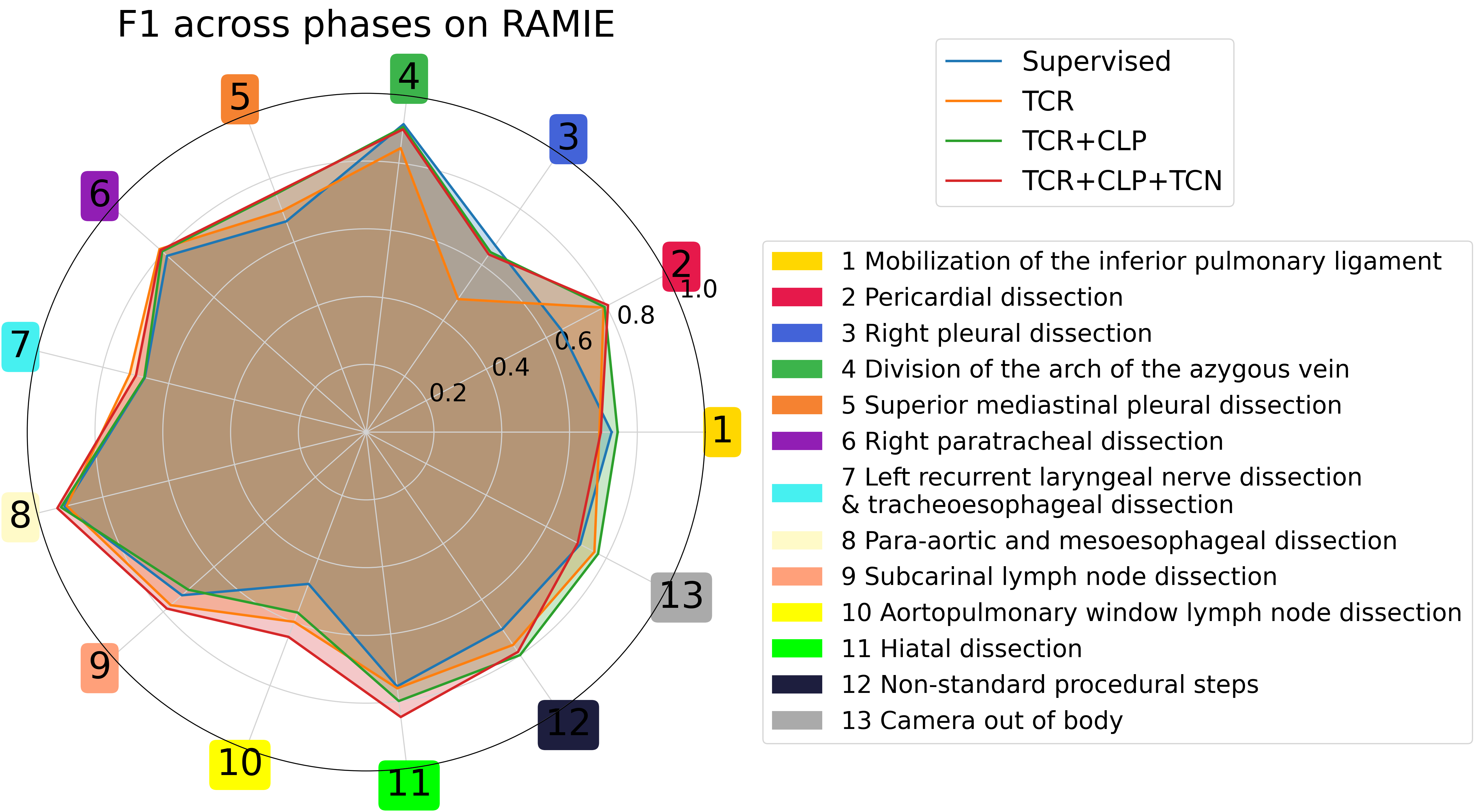}
    \caption{RAMIE: Qualitative result (left) and mean F1 scores across phases (right).}
    \label{fig:RAMIE}
\end{figure}

Our video transformer model can function independently by processing the current frame along with past frames, inherently capturing temporal dynamics.  TeCNO \cite{czempiel2020tecno}, widely used as a classification head in surgical phase recognition, is particularly common in self-supervised model evaluations. To assess its impact, we integrate TeCNO’s causal TCN, where our model functions as a spatio-temporal feature extractor. This distinguishes it from approaches that rely solely on frame-wise feature extraction. The results demonstrate that TeCNO further improves performance, as it excels at capturing long-term temporal dependencies. Results are shown in Table \ref{tab:ablation_study} and qualitative results and F1 across phases on both datasets can be found in Figure \ref{fig:Cholec80} and Figure \ref{fig:RAMIE}.

\subsubsection{State-of-the-art Comparison.} We compare our method to state-of-the-art approaches on both datasets. Given the limited research on semi-supervised learning for surgical phase recognition, we also evaluate self-supervised methods, including self-pretrained models with a temporal module based on TeCNO \cite{czempiel2020tecno}. The results are shown in Table \ref{comparison: RAMIE}. EndoFM \cite{wang2023foundation} was trained using DINO on over 33,000 endoscopic video clips, and SurgeNetXL \cite{jaspers2025scaling} was trained on 4,711,024 frames from 23 surgical procedures. Compared to these two large-scale self-pretraining approaches, our model outperforms both, achieving a 3\% and 0.9\% accuracy increase, respectively, while using significantly less surgery-specific unlabeled data. We also explored the use of additional unlabeled videos (beyond 20) on the RAMIE dataset, but did not observe further improvements in performance, which warrants further investigation. On the Cholec80 dataset, \cite{alapatt2024jumpstarting} reported an F1 score of 67.4 ± 4.9 using 5 labeled videos for training. Our method achieves comparable results. Endovit \cite{batic2024endovit} reported an accuracy of 84.68 ± 1.25 using 8 labeled videos, while our method performed similarly with only 5 labeled videos. Since these papers did not provide additional evaluation metrics, they are excluded from Table \ref{comparison:Cholec}. As shown in Table \ref{comparison:Cholec}, our method outperforms the semi-supervised SurgSSL \cite{shi2021semi} and matches the full supervision TeCNO performance on Cholec80, using only 1/4 of the labeled data.

\begin{table}[htbp!]
\centering
\caption{Comparison with state-of-the-art methods on RAMIE dataset. }
\label{comparison: RAMIE}
\resizebox{\textwidth}{!}{%
\begin{tabular}{lcccccc}
\hline
Method & \makecell{Supervision \\ Type}  & Accuracy (\%) & Precision (\%) & Recall (\%) & Jaccard (\%) \\
\hline
TeCNO \cite{czempiel2020tecno} & Fully supervised & 78.5 ± 4.0 & 73.9 ± 4.6 & 73.6 ± 5.1 & 58.3 ± 4.8 \\
FixMatch \cite{sohn2020fixmatch} & Semi-supervised & 79.2 ± 4.1 & 80.0 ± 4.1 & 75.5 ± 5.6 & 61.8 ± 4.8 \\
EndoFM \cite{wang2023foundation} & Self-supervised pretraining & 80.4 ± 3.3 & 79.0 ± 2.9 & 74.6 ± 5.6 & 61.9 ± 4.7 \\
SurgeNetXL \cite{jaspers2025scaling} & Self-supervised pretraining & 82.5 ± 3.7 & 79.4 ± 3.9 & 78.7 ± 3.9 & 64.8 ± 4.1 \\
Ours & Semi-supervised & \textbf{83.4 ± 3.7} & \textbf{81.6 ± 3.4} & \textbf{78.8 ± 3.8} & \textbf{66.1 ± 3.1} \\
\hline
\end{tabular}%
}

\end{table}
\begin{table}[htbp!]
\centering
\small 
\caption{Comparison with state-of-the-art methods on the Cholec80 dataset.}\label{comparison:Cholec}
\resizebox{\textwidth}{!}{

\begin{tabular}{lcccccc}
\hline
Method & \makecell{Supervision \\ Type} & \makecell{Labeled \\ Videos} & Accuracy (\%) & Precision (\%) & Recall (\%) & Jaccard (\%) \\
\hline
TeCNO \cite{czempiel2020tecno} & Fully supervised & 40 & 88.6 ± 7.8 & 86.5 ± 7.0 & 87.6 ± 6.7 & 75.1 ± 6.9 \\
\hline
SurgSSL \cite{shi2021semi} & \multirow{2}{*}{Semi-supervised} & \multirow{2}{*}{20} & 87.0 ± 7.4 & 84.2 ± 8.9 & 85.2 ± 11.1 & 70.5 ± 12.6 \\
Ours &&& \textbf{90.4 ± 7.0} & \textbf{88.4 ± 7.2} & \textbf{86.5 ± 7.4} & \textbf{77.5 ± 11.3} \\
\hline
SurgSSL \cite{shi2021semi} & \multirow{2}{*}{Semi-supervised} & \multirow{2}{*}{10} & 85.0 ± 7.7 & 83.3 ± 8.3 & 83.1 ± 12.3 & 68.0 ± 13.5 \\
Ours & & & \textbf{88.8 ± 6.5} & \textbf{85.0 ± 7.8} & \textbf{86.2 ± 6.1} & \textbf{74.5 ± 10.4} \\
\hline
SurgSSL \cite{shi2021semi} & \multirow{2}{*}{Semi-supervised} & \multirow{2}{*}{5} & 83.2 ± 7.7 & 81.8 ± 9.9 & 81.6 ± 12.3 & 65.6 ± 14.8 \\
Ours & & & \textbf{85.3 ± 7.6} & \textbf{84.2 ± 7.6} & \textbf{81.5 ± 7.8} & \textbf{69.2 ± 10.3} \\
\hline
\end{tabular}
}
\end{table}

\section{Conclusion}

In this work, we propose a semi-supervised framework for surgical phase recognition with a video transformer model. By incorporating long-short term temporal sampling and dynamic contrastive learning with class prototypes, our method effectively leverages unlabeled videos, demonstrating strong performance on both the Cholec80 and RAMIE datasets. Future work can explore better pseudo-label filtering to enhance performance on underrepresented classes, evaluate its effectiveness in federated learning, and analyze clinical relevance and failure cases for further improvements.

\subsubsection{Acknowledgments.}
This research was funded by Stichting Hanarth Fonds, study number: 2022-13. It is part of the INTRA-SURGE
(INTelligent computeR-Aided Surgical gUidance for Robot-assisted surGEry) project aimed at advancing the
future of surgery.

\subsubsection{Disclosure of Interests.}
 The authors have no competing interests to declare that
are relevant to the content of this article.

\bibliographystyle{splncs04}
\bibliography{references}





\end{document}